\documentclass[a4paper]{cas-sc}

\usepackage[numbers]{natbib}
\usepackage{graphicx}
\usepackage{float}
\usepackage{placeins}
\usepackage{caption}
\usepackage{amsmath}
\usepackage{amssymb}
\usepackage{placeins}
\usepackage{microtype}
\def\tsc#1{\csdef{#1}{\textsc{\lowercase{#1}}\xspace}}
\tsc{WGM}
\tsc{QE}
\tsc{EP}
\tsc{PMS}
\tsc{BEC}
\tsc{DE}

\begin{document}
\let\WriteBookmarks\relax
\def\floatpagepagefraction{1}
\def\textpagefraction{.001}
\let\printorcid\relax
\shorttitle{LGFN: Lightweight Gated RGB–Polarization Fusion with Modality-Availability Conditioning for Camouflaged Object Detection}
\shortauthors{Huang et al.}

\title[mode = title]{LGFN: Lightweight Gated RGB–Polarization Fusion with Modality-Availability Conditioning for Camouflaged Object Detection}

\author[address1]{Zhuangfan~Huang}
\ead{2112455033@stu.fosu.edu.cn}

\author[address1]{Xiaosong~Li}
\ead{lixiaosong@buaa.edu.cn}
\cormark[1]

\author[address1]{Yang~Liu}
\ead{ly25@fosu.edu.cn}

\author[address2]{Tao Ye}
 \ead{ayetao198715@163.com}

\author[address1]{Haishu~Tan}
\ead{tanhaishu@fosu.edu.cn}
\cormark[1]

\cortext[cor1]{Corresponding authors.}

\address[address1]{Guangdong-HongKong-Macao Joint Laboratory for Intelligent Micro-Nano Optoelectronic Technology, School of Physics and Optoelectronic Engineering, Foshan University, Foshan 528225, China.}

\address[address2]{School of Mechanical and Electrical Engineering, China University of Mining and Technology-Beijing, Beijing 100083, China.}


\begin{abstract}
Camouflaged object detection (COD) is an important engineering task in intelligent optical perception, but it remains challenging when targets closely resemble their surroundings. Polarization imaging provides complementary physical cues, whereas existing methods typically assume fixed multimodal input configurations and entangle intra-polarization coordination with interaction between red--green--blue (RGB) and polarization representations. We propose LGFN, a lightweight gated RGB--polarization fusion framework supporting separately optimized RGB-only and polarization-assisted configurations. A deterministic Modality Router selects the appropriate configuration according to polarization availability. In the multimodal configuration, an availability-conditioned Modality Gate calibrates the available polarization branches; the Gated Polarization Hub coordinates learned degree of linear polarization (DoLP) and angle of polarization (AoP) representations with explicit polarization cues; and RGB--Polarization Cross Fusion introduces the coordinated representation into the RGB hierarchy through controlled residual interaction. The multimodal configuration requires neither sample-dependent statistics nor handcrafted quality descriptors during inference. On the complete 230-image PCOD\_1200 test set, the RGB-only configuration achieves a mean absolute error of 0.0090, a Dice score of 0.8806, and an intersection over union of 0.8144, obtaining the best results on all six metrics among the evaluated RGB-based methods. Under a common local reevaluation protocol, the multimodal configuration outperforms PolarNet and IPNet on all six metrics. Relative to IPNet, it reduces the parameter count, floating-point operations, and latency by 53.1\%, 73.6\%, and 63.0\%, respectively. Component and objective ablations further support the effectiveness of the proposed polarization coordination and controlled cross-modal interaction.The source code will be available at  https://github.com/1hzf/LGFN.
\end{abstract}

\begin{keywords}
camouflaged object detection \sep polarization imaging \sep RGB--polarization fusion \sep polarization coordination \sep modality availability
\end{keywords}

\maketitle
\section{Introduction}
\label{sec:introduction}

Camouflaged object detection (COD) aims to segment objects whose appearance closely resembles that of their surroundings~\cite{fan2020camouflaged}. Camouflaged targets often exhibit weak foreground evidence, uncertain boundaries, and fragmented structures, making reliable localization particularly difficult. Existing methods based on red--green--blue (RGB) images improve contextual reasoning~\cite{mei2021pfnet}, multi-scale representation~\cite{pang2022zoomnet}, and structural modeling~\cite{ji2023dgnet}. However, they remain dependent on appearance-derived information. When foreground and background share similar color and texture characteristics, RGB observations alone may provide insufficient evidence for resolving their ambiguity.

Polarization imaging provides complementary physical cues related to surface material, reflection, and local geometry~\cite{yang2024polarimetricreview}. The degree of linear polarization (DoLP) and the angle of polarization (AoP) characterize different properties of reflected light and can reveal target--background discrepancies that are inconspicuous in RGB images. PolarNet demonstrated the feasibility of polarization-assisted COD~\cite{wang2023polarnet}, while IPNet introduced the PCOD\_1200 benchmark and a dual-flow RGB--polarization framework~\cite{wang2024ipnet}. More recently, a high-resolution adaptive fusion network further explored polarization-guided feature interaction for camouflage perception~\cite{wang2025hraf}. Nevertheless, existing methods generally rely on fixed multimodal inputs or strongly coupled cross-modal interaction. Directly combining RGB, DoLP, and AoP also requires one fusion module to handle both intra-polarization heterogeneity and RGB--polarization discrepancy, which may introduce unreliable polarization responses into the RGB representation.

To address these issues, we propose LGFN, a lightweight gated RGB--polarization fusion framework that separates polarization-domain coordination from cross-modal interaction. A deterministic Modality Router selects a separately optimized RGB-only or multimodal configuration according to polarization availability. Within the multimodal configuration, an availability-conditioned Modality Gate calibrates the available polarization branches without using sample-dependent statistics. The Gated Polarization Hub (GPH) coordinates learned DoLP and AoP representations with explicit polarization cues before cross-modal interaction. RGB--Polarization Cross Fusion (RPCF) then introduces the coordinated polarization representation into the RGB hierarchy through an attention-refined residual. This design separates intra-polarization coordination from RGB--polarization interaction while retaining RGB as the principal representation pathway. Fusion-consistency and modality-allocation objectives provide additional multimodal regularization without increasing inference-time computation.

The main contributions are summarized as follows:
\begin{itemize}
    \item We propose LGFN, a lightweight gated RGB--polarization fusion framework that supports separately optimized RGB-only and polarization-assisted configurations. A deterministic Modality Router selects the appropriate configuration according to polarization availability.

    \item We develop an availability-conditioned Modality Gate that uses only the modality-availability vector to allocate the available DoLP and AoP branches and regulate the overall strength of polarization injection, without relying on sample-dependent quality estimates during inference.

    \item We introduce GPH and RPCF to separate intra-polarization coordination from RGB--polarization interaction. GPH coordinates learned and explicit polarization cues, whereas RPCF introduces the coordinated representation into the RGB hierarchy through controlled residual fusion.
\end{itemize}

\section{Related Work}
\label{sec:related_work}

Research related to this work spans three closely connected directions. RGB-based COD improves visual discrimination within the appearance domain, polarization-assisted COD introduces complementary physical evidence, and multimodal learning investigates how heterogeneous information can be integrated under different input conditions. We review these directions in sequence and clarify the remaining gap addressed by LGFN.

\subsection{RGB-Based COD}
\label{sec:rw_rgb_cod}

Early RGB-based COD methods mainly improve the distinction between camouflaged objects and their surroundings through contextual reasoning and hierarchical aggregation. The COD10K study established a large-scale benchmark and introduced a search-and-identification framework for concealed-object segmentation~\cite{fan2020camouflaged}. PFNet suppresses distracting responses through distraction mining, while C2FNet strengthens contextual representation through cross-level fusion~\cite{mei2021pfnet,sun2021c2fnet}. Graph interaction and joint localization, segmentation, and ranking have also been explored to improve target discovery and multi-object reasoning~\cite{zhai2021mgl,lv2021simultaneously}.

Subsequent studies increasingly focus on ambiguity modeling, scale variation, and structural recovery. UGTR introduces uncertainty-guided Transformer reasoning, whereas ZoomNet and SegMaR improve difficult-target localization through mixed-scale processing and iterative refinement~\cite{yang2021ugtr,pang2022zoomnet,jia2022segmar}. Hierarchical Transformer representation and high-resolution feedback are further investigated in FSPNet and HitNet~\cite{huang2023fspnet,hu2023hitnet}. Other methods exploit gradient, frequency, boundary, and foreground-context information~\cite{ji2023dgnet,he2023feder,yin2024camoformer}. Boundary-guided refinement and frequency-domain edge perception have also been explored to recover ambiguous camouflage structures~\cite{xu2021boundary,fang2025epfdnet}. Recent studies further explore contrastive representation learning, knowledge-guided collaboration, boundary localization, visual prompts, mask guidance, dual-path pyramids, boundary-aware gating, and geometric priors~\cite{guo2025clad,wu2025knowledge,zhang2025boundary,li2025weakly,luo2024vscode,he2025hdpnet,wang2026gbnet,han2026depthsam}. Although these approaches substantially strengthen RGB representations, they still derive their evidence mainly from visual appearance. This limitation motivates the use of sensing modalities that provide physically distinct information beyond RGB observations.

\subsection{Polarization-Assisted COD}
\label{sec:rw_polar_cod}

Motivated by the ambiguity of appearance-based evidence, polarization imaging provides an alternative means of distinguishing visually similar regions. Reflected polarization is influenced by surface material, roughness, geometry, and illumination, allowing it to reveal differences that may remain inconspicuous in RGB observations~\cite{tyo2006polarimetry}. Polarimetric information has therefore been explored for material perception, surface analysis, depth estimation, and three-dimensional reconstruction~\cite{yang2024polarimetricreview,taglione2024polarimetric}. DoLP describes polarization strength, whereas AoP characterizes its dominant orientation, providing complementary physical cues for camouflage perception.

PolarNet demonstrated the feasibility of polarization-assisted COD~\cite{wang2023polarnet}. IPNet subsequently introduced the PCOD\_1200 dataset and employed a dual-flow architecture for RGB--polarization interaction~\cite{wang2024ipnet}. More recently, a high-resolution adaptive fusion network further strengthened polarization-assisted camouflage perception through adaptive feature interaction~\cite{wang2025hraf}. These studies establish the value of polarization information, but mainly assume fixed multimodal inputs and focus on direct cross-modal interaction. Consequently, heterogeneity between DoLP and AoP must often be resolved together with the discrepancy between polarization and RGB representations. This shifts the central question from whether polarization is useful to how heterogeneous polarization cues should be coordinated and introduced without disturbing reliable RGB features.
\begin{figure}[pos=t]
    \centering
    \includegraphics[width=\textwidth]{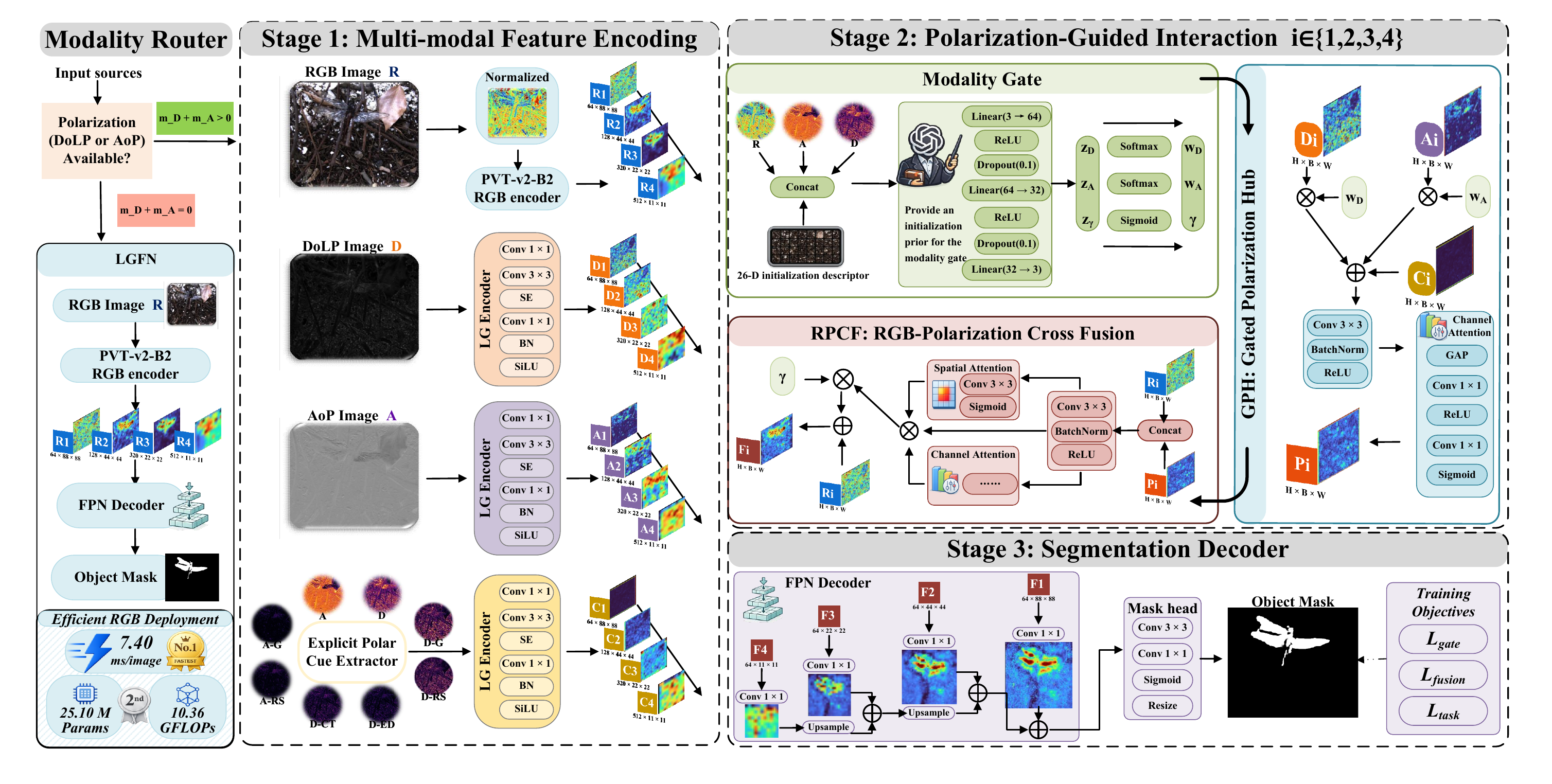}
    \caption{Overall architecture of LGFN. The deterministic Modality Router selects the RGB-only or multimodal route according to polarization availability. GPH coordinates heterogeneous polarization cues, and RPCF performs controlled polarization injection into the RGB hierarchy. Solid and dashed arrows denote inference paths and training-only supervision, respectively.}
    \label{fig:lgfn_overall}
\end{figure}
\subsection{Multimodal Fusion and Routing}
\label{sec:rw_multimodal}

This question connects polarization-assisted COD to the broader problem of multimodal fusion. EMMA develops an equivariant self-supervised framework for fusion without ground-truth fused images~\cite{zhao2024emma}, while TC-MoA employs task-customized adapters to support different fusion requirements within a unified model~\cite{zhu2024tcmoa}. SHIP further investigates local and global high-order interactions between modalities~\cite{zhou2025ship}. Progressive token exchange and semantic-knowledge feedback have also been explored to strengthen cross-modal interaction and improve the downstream utility of fused representations~\cite{huang2024ptet,zhou2025skffusion}. These studies demonstrate the importance of adaptive cross-modal interaction. However, general image-fusion methods primarily aim to synthesize a new image, whereas COD requires task-oriented feature interaction that preserves reliable semantic representations. For polarization-assisted COD, this suggests an asymmetric strategy in which RGB remains the principal pathway and polarization provides controlled complementary evidence.

Practical multimodal systems must also accommodate variations in sensor availability. ShaSpec learns shared and modality-specific features for downstream tasks under missing-modality conditions~\cite{wang2023shaspec}. Existing missing-modality approaches commonly accommodate multiple input combinations within a shared network, which may retain modality-specific branches even when the corresponding inputs are unavailable. For polarization-assisted COD, however, practical deployment additionally requires the inference process to adapt efficiently to different RGB and polarization input configurations. Taken together, three requirements remain difficult to address jointly: coordinating heterogeneous polarization cues before cross-modal interaction, preserving the RGB hierarchy during polarization injection, and adapting the inference pathway to polarization availability without unnecessary computation. These considerations motivate the LGFN framework introduced in the next section.

\section{Proposed Method}
\label{sec:method}

\subsection{Overall Framework}
\label{sec:overall}

Motivated by the limitations discussed above, LGFN is designed to coordinate heterogeneous polarization cues, preserve reliable RGB representations during cross-modal interaction, and adapt its inference pathway to polarization availability. Given a mandatory RGB image $\mathbf{I}\in\mathbb{R}^{3\times H\times W}$ and optional polarization inputs including a DoLP map $\mathbf{D}\in\mathbb{R}^{1\times H\times W}$ and an AoP map $\mathbf{A}\in\mathbb{R}^{1\times H\times W}$, LGFN predicts a camouflage probability map $\widehat{\mathbf{Y}}\in[0,1]^{H\times W}$. Rather than treating DoLP and AoP as additional appearance channels, LGFN first coordinates the available polarization information within the polarization domain and then introduces the resulting representation into an RGB-centered segmentation pathway.

As illustrated in Fig.~\ref{fig:lgfn_overall}, LGFN contains two independently optimized inference routes selected according to polarization availability. Let $\mathbf{m}=[1,m_D,m_A]$ denote the modality-availability vector, where RGB is mandatory and $m_D,m_A\in\{0,1\}$ indicate the availability of DoLP and AoP, respectively. When $m_D+m_A=0$, LGFN activates the RGB-only route; when $m_D+m_A>0$, the presence of at least one polarization input activates the multimodal route. The routing rule is defined as
\begin{equation}
\widehat{\mathbf{Y}}
=
\begin{cases}
\mathcal{F}_{\mathrm{RGB}}
\left(
\mathbf{I};
\boldsymbol{\Theta}_{\mathrm{RGB}}
\right),
& m_D+m_A=0,
\\[3pt]
\mathcal{F}_{\mathrm{MM}}
\left(
\mathbf{I},
m_D\mathbf{D},
m_A\mathbf{A},
\mathbf{m};
\boldsymbol{\Theta}_{\mathrm{MM}}
\right),
& m_D+m_A>0.
\end{cases}
\label{eq:system_routing}
\end{equation}

Here, $\mathcal{F}_{\mathrm{RGB}}$ and $\mathcal{F}_{\mathrm{MM}}$ denote the RGB-only and multimodal inference routes of LGFN, respectively, while $\boldsymbol{\Theta}_{\mathrm{RGB}}$ and $\boldsymbol{\Theta}_{\mathrm{MM}}$ denote their independently optimized parameters. The deterministic Modality Router depends only on polarization availability and does not estimate sample-specific input quality or combine predictions from multiple routes. Consequently, when polarization inputs are unavailable, LGFN directly activates the dedicated RGB-only route without retaining inactive polarization-specific components.

Since the RGB-only route follows a conventional RGB segmentation pipeline, the remainder of this section focuses on the multimodal route of LGFN. The RGB image is encoded by a PVT-v2-B2 backbone~\cite{wang2022pvtv2}, while DoLP and AoP are processed by two separate lightweight encoders with unshared parameters. The three encoders produce four-level representations $\{\mathbf{R}_i\}_{i=1}^{4}$, $\{\mathbf{D}_i\}_{i=1}^{4}$, and $\{\mathbf{A}_i\}_{i=1}^{4}$. In parallel, an explicit cue encoder extracts scale-aligned polarization structures $\{\mathbf{C}_i\}_{i=1}^{4}$ to retain local boundary and contrast information that may be weakened during hierarchical encoding.

Within the multimodal route, cross-modal interaction is organized into three successive stages. The Modality Gate first determines availability-conditioned polarization coefficients, GPH then coordinates DoLP, AoP, and explicit polarization cues within the polarization domain, and RPCF finally introduces the coordinated polarization representation into the RGB hierarchy through controlled residual interaction. The multimodal pipeline is summarized as
\begin{equation}
\small
\begin{gathered}
(w_D,w_A,\gamma)
=
\mathcal{G}(\mathbf{m}),
\\
\mathbf{P}_i
=
\mathcal{H}_i
(\mathbf{D}_i,\mathbf{A}_i,\mathbf{C}_i;w_D,w_A,\mathbf{m}),
\qquad
\mathbf{F}_i
=
\mathcal{X}_i
(\mathbf{R}_i,\mathbf{P}_i;\gamma),
\\
\widehat{\mathbf{Y}}
=
\sigma
\left\{
\operatorname{Up}
\left[
\mathcal{D}
\left(
\{\mathbf{F}_i\}_{i=1}^{4}
\right)
\right]
\right\},
\qquad
i\in\{1,2,3,4\}.
\end{gathered}
\label{eq:overall_mapping}
\end{equation}

In Eq.~\eqref{eq:overall_mapping}, $\mathcal{G}$ denotes the Modality Gate, while $\mathcal{H}_i$ and $\mathcal{X}_i$ denote GPH and RPCF at the $i$-th scale, respectively. The coefficients $w_D$ and $w_A$ regulate the contributions of the available polarization branches, whereas $\gamma$ controls the overall injection strength. The coordinated polarization feature $\mathbf{P}_i$ is introduced into the RGB feature $\mathbf{R}_i$ to produce $\mathbf{F}_i$, which is subsequently aggregated by a Feature Pyramid Network (FPN)-style decoder $\mathcal{D}$~\cite{lin2017fpn}. Finally, $\operatorname{Up}(\cdot)$ restores the prediction to the input resolution and $\sigma(\cdot)$ produces the camouflage probability map. The following subsections detail the three core components.
\subsection{Modality Gate}
\label{sec:gate}

Following the system-level routing decision, the multimodal route further calibrates the contributions of the available polarization branches. DoLP and AoP provide complementary physical information, while their availability may vary across sensing configurations. A fixed polarization allocation cannot explicitly adapt to such changes in input availability. We therefore introduce an availability-conditioned Modality Gate that operates solely on the modality vector $\mathbf{m}=[1,m_D,m_A]$, where $m_D,m_A\in\{0,1\}$ indicate the presence of DoLP and AoP, respectively. The Modality Gate receives no image content, handcrafted statistics, or sample-level quality descriptors.

\begin{figure}[pos=t]
\centering
\includegraphics[width=\linewidth]{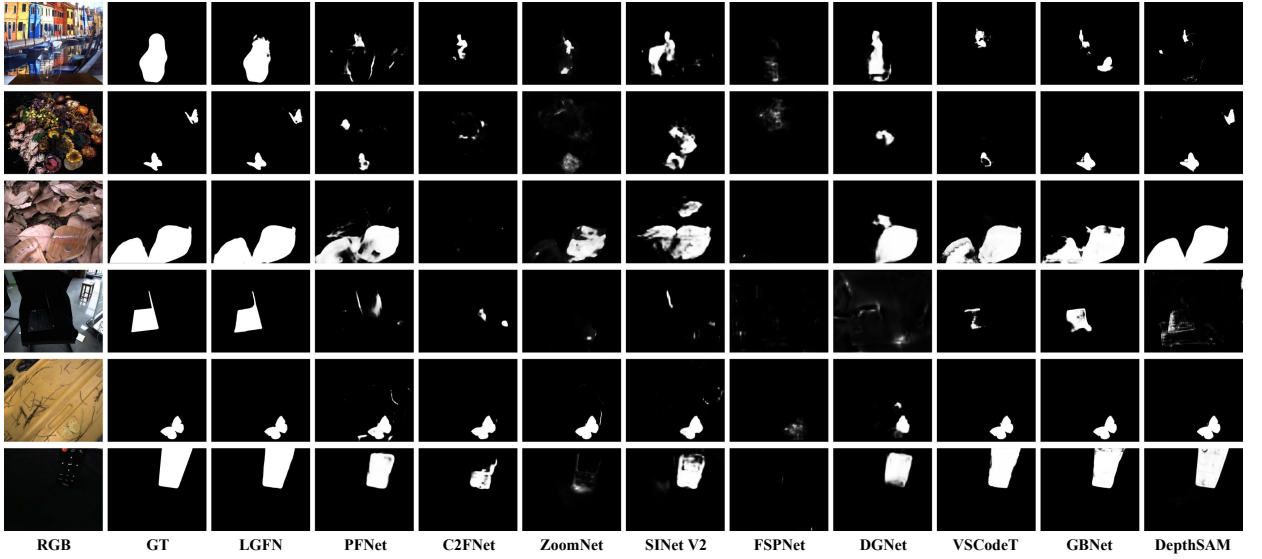}
\caption{Qualitative comparison with RGB-based camouflaged object detection methods on representative samples from the PCOD\_1200 test set.}
\label{fig:pcod_rgb_visual}
\end{figure}

A lightweight multilayer perceptron maps $\mathbf{m}$ to three logits $[z_D,z_A,z_{\gamma}]$, which control the relative allocation of DoLP and AoP and the overall polarization-injection strength. The polarization weights are obtained by a masked softmax:
\begin{equation}
w_k
=
\frac{
m_k\exp(z_k)
}{
\displaystyle
\sum_{j\in\{D,A\}}
m_j\exp(z_j)
},
\qquad
k\in\{D,A\}.
\label{eq:gate_weights}
\end{equation}

In Eq.~\eqref{eq:gate_weights}, an unavailable polarization branch is assigned zero weight, while the available branches are normalized such that $w_D+w_A=1$ when $m_D+m_A>0$. Thus, $w_D$ and $w_A$ represent availability-conditioned polarization allocation rather than sample-specific quality estimates.

The Gate additionally predicts a bounded coefficient $\gamma$ to control the overall strength of polarization injection:
\begin{equation}
\gamma
=
\eta_{\mathrm{pol}}
\left[
0.3
+
0.7
\sigma
\left(
z_{\gamma}
\right)
\right],
\qquad
\eta_{\mathrm{pol}}
=
\mathbf{1}
\left[
m_D+m_A>0
\right].
\label{eq:gate_gamma}
\end{equation}

Here, $\sigma(\cdot)$ denotes the sigmoid function and $\eta_{\mathrm{pol}}$ indicates whether polarization information is available. When $m_D+m_A>0$, $\gamma\in[0.3,1]$ regulates the overall contribution of the coordinated polarization representation. The case $\gamma=0$ when $m_D+m_A=0$ is retained only as a mathematical fallback, since RGB-only inference is handled directly by the dedicated RGB-only route. Together, $w_D$, $w_A$, and $\gamma$ provide availability-conditioned global calibration: $w_D$ and $w_A$ regulate polarization aggregation in GPH, while $\gamma$ controls the subsequent residual injection in RPCF.

\subsection{Gated Polarization Hub}
\label{sec:gph}

Building on the allocation coefficients produced by the Modality Gate, GPH coordinates heterogeneous polarization evidence before cross-modal interaction. Although DoLP and AoP provide complementary physical cues, they exhibit distinct response characteristics and representation semantics. Directly introducing them into the RGB hierarchy would require the cross-modal module to simultaneously handle intra-polarization heterogeneity and RGB--polarization discrepancy. GPH therefore first coordinates the available DoLP and AoP representations within the polarization domain to construct a coordinated polarization representation.

In addition to the learned DoLP and AoP features, GPH introduces explicit polarization cues to preserve local structural information that may be weakened during hierarchical encoding. Unavailable source maps are first suppressed according to $m_D$ and $m_A$. An eight-channel cue tensor is then constructed from the masked DoLP and AoP maps, their first-order gradient magnitudes, $5\times5$ local residuals, and the maximum gradient and residual responses across the available polarization modalities. A lightweight cue encoder transforms these source-level cues into four scale-aligned representations $\{\mathbf{C}_i\}_{i=1}^{4}$ for subsequent polarization coordination.

At the $i$-th scale, GPH combines the available DoLP and AoP features with the corresponding explicit polarization cues:
\begin{equation}
\mathbf{U}_{i}
=
w_D\mathbf{D}_{i}
+
w_A\mathbf{A}_{i}
+
\mathbf{C}_{i},
\qquad
i\in\{1,2,3,4\}.
\label{eq:gph_aggregation}
\end{equation}

Here, $\mathbf{D}_{i}$, $\mathbf{A}_{i}$, and $\mathbf{C}_{i}$ denote the DoLP, AoP, and explicit cue features at the $i$-th scale, respectively, and $\mathbf{U}_{i}$ is the resulting polarization representation. The masked weights $w_D$ and $w_A$ suppress unavailable learned branches, while $\mathbf{C}_{i}$ provides complementary local structural evidence extracted from the available polarization inputs.

The aggregated representation is further refined through a scale-specific convolutional projection:
\begin{equation}
\mathbf{V}_{i}
=
\delta
\left[
\operatorname{BN}_{i}
\left(
\operatorname{Conv}^{i}_{3\times3}
\left(
\mathbf{U}_{i}
\right)
\right)
\right].
\label{eq:gph_refinement}
\end{equation}

Here, $\operatorname{Conv}^{i}_{3\times3}(\cdot)$, $\operatorname{BN}_{i}(\cdot)$, and $\delta(\cdot)$ denote scale-specific convolution, batch normalization, and rectified linear unit (ReLU) activation, respectively. Each scale employs independent parameters to adapt the polarization representation to its corresponding feature level.

\begin{table*}[!t]
\centering
\footnotesize
\renewcommand{\arraystretch}{1.0}
{\rmfamily
\caption{Quantitative comparison with RGB-based camouflaged object detection methods on the PCOD\_1200 test set. The best, second-best, and third-best results are highlighted in red, blue, and green, respectively.}
\label{tab:rgb_comparison}

\begin{tabular*}{\textwidth}{
@{\extracolsep{\fill}}
ccccccccc
@{}
}
\toprule
\multirow{2}{*}{Methods}
& \multirow{2}{*}{Year}
& \multirow{2}{*}{Venue}
& \multicolumn{6}{c}{Evaluation metrics} \\
\cmidrule(lr){4-9}
& & &
MAE$\downarrow$
& $S_{\alpha}\uparrow$
& $E_{\phi}\uparrow$
& $F_{\beta}^{w}\uparrow$
& Dice$\uparrow$
& IoU$\uparrow$ \\
\midrule

PFNet~\cite{mei2021pfnet}
& 2021 & CVPR
& 0.0375 & 0.7596 & 0.8215 & 0.6220 & 0.6244 & 0.5256 \\

C2FNet~\cite{sun2021c2fnet}
& 2021 & IJCAI
& 0.0308 & 0.7727 & 0.8394 & 0.6319 & 0.6336 & 0.5441 \\

ZoomNet~\cite{pang2022zoomnet}
& 2022 & CVPR
& 0.0340 & 0.7546 & 0.7883 & 0.5904 & 0.5845 & 0.5122 \\

SINet-V2~\cite{fan2022sinetv2}
& 2022 & TPAMI
& 0.0359 & 0.7686 & 0.8368 & 0.6202 & 0.6347 & 0.5433 \\

FSPNet~\cite{huang2023fspnet}
& 2023 & CVPR
& 0.0498 & 0.5341 & 0.4053 & 0.1876 & 0.0691 & 0.0383 \\

DGNet~\cite{ji2023dgnet}
& 2023 & MIR
& 0.0289 & 0.7913 & 0.8585 & 0.6493 & 0.6639 & 0.5728 \\

VSCode-T~\cite{luo2024vscode}
& 2024 & CVPR
& 0.0219 & 0.8239 & 0.8722 & 0.7141 & 0.7213 & 0.6411 \\

GBNet~\cite{wang2026gbnet}
& 2026 & TIP
& \textbf{\color[HTML]{00A651}0.0158}
& \textbf{\color[HTML]{00A651}0.8840}
& \textbf{\color[HTML]{34CDF9}0.9246}
& \textbf{\color[HTML]{34CDF9}0.8152}
& \textbf{\color[HTML]{34CDF9}0.8249}
& \textbf{\color[HTML]{00A651}0.7651} \\

DepthSAM~\cite{han2026depthsam}
& 2026 & CVPR
& \textbf{\color[HTML]{34CDF9}0.0140}
& \textbf{\color[HTML]{34CDF9}0.8870}
& \textbf{\color[HTML]{00A651}0.9213}
& \textbf{\color[HTML]{00A651}0.8149}
& \textbf{\color[HTML]{00A651}0.8245}
& \textbf{\color[HTML]{34CDF9}0.7714} \\

\midrule

\textbf{LGFN }
& 2026 & --
& \textbf{\color[HTML]{FE0000}0.0090}
& \textbf{\color[HTML]{FE0000}0.9162}
& \textbf{\color[HTML]{FE0000}0.9637}
& \textbf{\color[HTML]{FE0000}0.8688}
& \textbf{\color[HTML]{FE0000}0.8806}
& \textbf{\color[HTML]{FE0000}0.8144} \\

\bottomrule
\end{tabular*}
}
\end{table*}

\begin{figure}[pos=t]
\centering
\includegraphics[width=\linewidth]{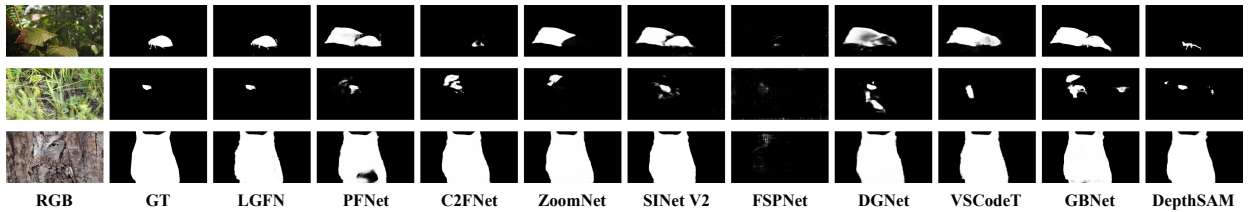}
\caption{Qualitative comparison with RGB-based camouflaged object detection methods on representative samples from the fixed 200-image NC4K evaluation subset.}
\label{fig:nc4k_rgb_visual}
\end{figure}

GPH further applies SE-style channel recalibration~\cite{hu2018senet} to emphasize informative polarization responses and suppress redundant channels:
\begin{equation}
\begin{gathered}
\boldsymbol{\alpha}_{i}
=
\sigma
\left[
\mathbf{W}^{i}_{2}
\delta
\left(
\mathbf{W}^{i}_{1}
\operatorname{GAP}
\left(
\mathbf{V}_{i}
\right)
\right)
\right],
\\[3pt]
\mathbf{P}_{i}
=
\eta_{\mathrm{pol}}
\left(
\boldsymbol{\alpha}_{i}
\odot
\mathbf{V}_{i}
\right),
\qquad
i\in\{1,2,3,4\}.
\end{gathered}
\label{eq:gph_output}
\end{equation}

In Eq.~\eqref{eq:gph_output}, $\operatorname{GAP}(\cdot)$ denotes global average pooling, $\mathbf{W}^{i}_{1}$ and $\mathbf{W}^{i}_{2}$ are channel-reduction and restoration projections, and $\boldsymbol{\alpha}_{i}$ denotes the channel-wise recalibration weights. The polarization-availability indicator $\eta_{\mathrm{pol}}=\mathbf{1}[m_D+m_A>0]$ provides a mathematical suppression mechanism when no polarization input is present. The resulting $\mathbf{P}_{i}$ is the coordinated polarization feature produced by GPH at the $i$-th scale.

By coordinating heterogeneous polarization evidence before cross-modal interaction, GPH reduces the direct influence of uncoordinated DoLP and AoP responses on the RGB hierarchy. The resulting features $\{\mathbf{P}_{i}\}_{i=1}^{4}$ are then delivered to RPCF for controlled residual interaction with the corresponding RGB features.

\subsection{RGB--Polarization Cross Fusion}
\label{sec:rpcf}

Following polarization-domain coordination in GPH, RPCF introduces the coordinated polarization representation into the RGB hierarchy through asymmetric residual interaction. Rather than symmetrically mixing the two modalities, RPCF preserves RGB as the principal representation pathway and treats polarization as controlled complementary evidence. The same interaction form is applied independently at all four feature scales with scale-specific learnable parameters.

At the $i$-th scale, the RGB feature $\mathbf{R}_i$ and coordinated polarization feature $\mathbf{P}_i$ are concatenated and projected to construct a polarization-conditioned residual candidate:
\begin{equation}
\boldsymbol{\Delta}_{i}
=
\phi_i
\left(
\operatorname{Cat}
\left(
\mathbf{R}_{i},
\mathbf{P}_{i}
\right)
\right),
\qquad
i\in\{1,2,3,4\}.
\label{eq:rpcf_increment}
\end{equation}

Here, $\operatorname{Cat}(\cdot)$ denotes channel-wise concatenation, and $\phi_i(\cdot)$ is a scale-specific $1\times1$ projection that restores the channel dimension of $\mathbf{R}_i$. The resulting $\boldsymbol{\Delta}_{i}$ serves as the cross-modal residual candidate for subsequent attention-based recalibration.

\begin{table*}[!t]
\centering
\footnotesize
\renewcommand{\arraystretch}{1.0}
{\rmfamily
\caption{Quantitative comparison with RGB-based camouflaged object detection methods on the fixed 200-image NC4K evaluation subset. }
\label{tab:nc4k_comparison}

\begin{tabular*}{\textwidth}{
@{\extracolsep{\fill}}
ccccccccc
@{}
}
\toprule
\multirow{2}{*}{Methods}
& \multirow{2}{*}{Year}
& \multirow{2}{*}{Venue}
& \multicolumn{6}{c}{Evaluation metrics} \\
\cmidrule(lr){4-9}
& & &
MAE$\downarrow$
& $S_{\alpha}\uparrow$
& $E_{\phi}\uparrow$
& $F_{\beta}^{w}\uparrow$
& Dice$\uparrow$
& IoU$\uparrow$ \\
\midrule

PFNet~\cite{mei2021pfnet}
& 2021 & CVPR
& 0.0276 & 0.9281 & 0.9559 & 0.9190 & 0.9220 & 0.8699 \\

C2FNet~\cite{sun2021c2fnet}
& 2021 & IJCAI
& 0.0234 & 0.9328 & 0.9596 & 0.9233 & 0.9265 & 0.8754 \\

ZoomNet~\cite{pang2022zoomnet}
& 2022 & CVPR
& 0.0204 & 0.9411 & 0.9648 & 0.9316 & 0.9351 & 0.8933 \\

SINet-V2~\cite{fan2022sinetv2}
& 2022 & TPAMI
& 0.0242 & 0.9341 & 0.9617 & 0.9256 & 0.9297 & 0.8789 \\

FSPNet~\cite{huang2023fspnet}
& 2023 & CVPR
& 0.1487 & 0.4893 & 0.3936 & 0.2744 & 0.1412 & 0.0907 \\

DGNet~\cite{ji2023dgnet}
& 2023 & MIR
& 0.0208 & 0.9380 & 0.9647 & 0.9260 & 0.9332 & 0.8861 \\

VSCode-T~\cite{luo2024vscode}
& 2024 & CVPR
& 0.0159 & 0.9529 & 0.9732 & 0.9428 & 0.9474 & 0.9112 \\

GBNet~\cite{wang2026gbnet}
& 2026 & TIP
& \textbf{\color[HTML]{00A651}0.0109}
& \textbf{\color[HTML]{00A651}0.9607}
& \textbf{\color[HTML]{00A651}0.9812}
& \textbf{\color[HTML]{34CDF9}0.9613}
& \textbf{\color[HTML]{00A651}0.9624}
& \textbf{\color[HTML]{00A651}0.9325} \\

DepthSAM~\cite{han2026depthsam}
& 2026 & CVPR
& \textbf{\color[HTML]{FE0000}0.0092}
& \textbf{\color[HTML]{FE0000}0.9673}
& \textbf{\color[HTML]{34CDF9}0.9840}
& \textbf{\color[HTML]{FE0000}0.9679}
& \textbf{\color[HTML]{34CDF9}0.9684}
& \textbf{\color[HTML]{FE0000}0.9431} \\

\midrule

\textbf{LGFN }
& 2026 & --
& \textbf{\color[HTML]{34CDF9}0.0096}
& \textbf{\color[HTML]{34CDF9}0.9654}
& \textbf{\color[HTML]{FE0000}0.9874}
& \textbf{\color[HTML]{00A651}0.9611}
& \textbf{\color[HTML]{FE0000}0.9688}
& \textbf{\color[HTML]{34CDF9}0.9398} \\

\bottomrule
\end{tabular*}
}
\end{table*}
\begin{figure}[pos=t]
\centering
\includegraphics[width=\linewidth]{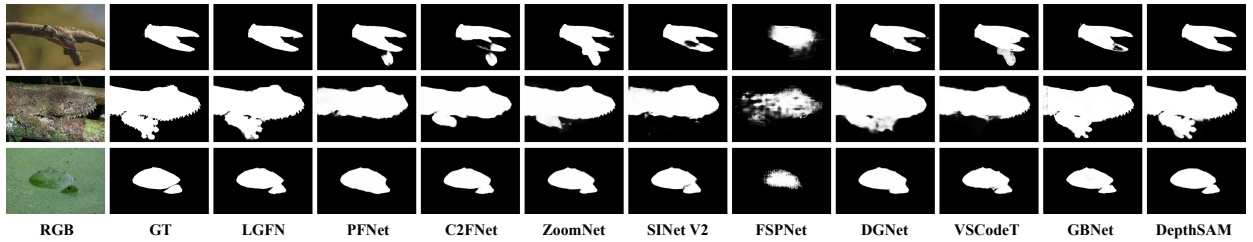}
\caption{Qualitative comparison with RGB-based camouflaged object detection methods on representative samples from the fixed 405-image COD10K evaluation subset.}
\label{fig:cod10k_rgb_visual}
\end{figure}

The residual candidate is further recalibrated by parallel spatial- and channel-attention branches, which respectively emphasize informative regions and cross-modal channels. The attention-refined residual and the fused RGB representation are defined as
\begin{equation}
\begin{gathered}
\widehat{\boldsymbol{\Delta}}_{i}
=
\mathcal{S}_{i}
\left(
\boldsymbol{\Delta}_{i}
\right)
\odot
\mathcal{C}_{i}
\left(
\boldsymbol{\Delta}_{i}
\right)
\odot
\boldsymbol{\Delta}_{i},
\\[3pt]
\mathbf{F}_{i}
=
\mathbf{R}_{i}
+
1.5\gamma
\widehat{\boldsymbol{\Delta}}_{i}.
\end{gathered}
\label{eq:rpcf_fusion}
\end{equation}

In Eq.~\eqref{eq:rpcf_fusion}, $\mathcal{S}_{i}(\cdot)$ and $\mathcal{C}_{i}(\cdot)$ denote the spatial- and channel-attention operations at the $i$-th scale, respectively, $\odot$ denotes element-wise multiplication with broadcasting, and $\widehat{\boldsymbol{\Delta}}_{i}$ is the attention-refined polarization residual. The coefficient $\gamma$ is produced by the Modality Gate and controls the overall injection strength, while the residual scale is empirically fixed at $1.5$ for all experiments. Through the identity shortcut, $\mathbf{F}_{i}$ retains the original RGB representation and introduces only the calibrated polarization increment.

This asymmetric residual interaction allows polarization evidence to complement ambiguous target structures while preserving the RGB hierarchy as the principal representation pathway. When no polarization modality is available, $\gamma=0$ formally reduces RPCF to $\mathbf{F}_{i}=\mathbf{R}_{i}$; this case serves only as a mathematical fallback, since RGB-only inference is handled by the dedicated RGB-only route of LGFN. The four fused representations $\{\mathbf{F}_{i}\}_{i=1}^{4}$ are finally aggregated by the multimodal FPN decoder to produce the camouflage probability map.

\begin{table*}[!t]
\centering
\footnotesize
\renewcommand{\arraystretch}{1.0}
{\rmfamily
\caption{Quantitative comparison with RGB-based camouflaged object detection methods on the fixed 405-image COD10K evaluation subset.}
\label{tab:cod10k_comparison}

\begin{tabular*}{\textwidth}{
@{\extracolsep{\fill}}
ccccccccc
@{}
}
\toprule
\multirow{2}{*}{Methods}
& \multirow{2}{*}{Year}
& \multirow{2}{*}{Venue}
& \multicolumn{6}{c}{Evaluation metrics} \\
\cmidrule(lr){4-9}
& & &
MAE$\downarrow$
& $S_{\alpha}\uparrow$
& $E_{\phi}\uparrow$
& $F_{\beta}^{w}\uparrow$
& Dice$\uparrow$
& IoU$\uparrow$ \\
\midrule

PFNet~\cite{mei2021pfnet}
& 2021 & CVPR
& 0.0427 & 0.8056 & 0.8789 & 0.7132 & 0.7237 & 0.6165 \\

C2FNet~\cite{sun2021c2fnet}
& 2021 & IJCAI
& 0.0400 & 0.8164 & 0.8846 & 0.7244 & 0.7331 & 0.6314 \\

ZoomNet~\cite{pang2022zoomnet}
& 2022 & CVPR
& 0.0348 & 0.8353 & 0.8853 & 0.7619 & 0.7599 & 0.6708 \\

SINet-V2~\cite{fan2022sinetv2}
& 2022 & TPAMI
& 0.0388 & 0.8191 & 0.8847 & 0.7265 & 0.7403 & 0.6364 \\

FSPNet~\cite{huang2023fspnet}
& 2023 & CVPR
& 0.0692 & 0.5893 & 0.5056 & 0.3465 & 0.2464 & 0.1777 \\

DGNet~\cite{ji2023dgnet}
& 2023 & MIR
& 0.0347 & 0.8321 & 0.9058 & 0.7442 & 0.7629 & 0.6588 \\

VSCode-T~\cite{luo2024vscode}
& 2024 & CVPR
& 0.0308
& 0.8497
& 0.9158
& \textbf{\color[HTML]{00A651}0.7892}
& 0.7906
& 0.6976 \\

GBNet~\cite{wang2026gbnet}
& 2026 & TIP
& \textbf{\color[HTML]{34CDF9}0.0216}
& \textbf{\color[HTML]{34CDF9}0.8985}
& \textbf{\color[HTML]{34CDF9}0.9500}
& \textbf{\color[HTML]{34CDF9}0.8552}
& \textbf{\color[HTML]{34CDF9}0.8636}
& \textbf{\color[HTML]{34CDF9}0.7915} \\

DepthSAM~\cite{han2026depthsam}
& 2026 & CVPR
& \textbf{\color[HTML]{FE0000}0.0173}
& \textbf{\color[HTML]{FE0000}0.9141}
& \textbf{\color[HTML]{FE0000}0.9568}
& \textbf{\color[HTML]{FE0000}0.8759}
& \textbf{\color[HTML]{FE0000}0.8850}
& \textbf{\color[HTML]{FE0000}0.8201} \\

\midrule

\textbf{LGFN }
& 2026 & --
& \textbf{\color[HTML]{00A651}0.0282}
& \textbf{\color[HTML]{00A651}0.8540}
& \textbf{\color[HTML]{00A651}0.9179}
& 0.7820
& \textbf{\color[HTML]{00A651}0.7966}
& \textbf{\color[HTML]{00A651}0.7068} \\

\bottomrule
\end{tabular*}
}
\end{table*}

\begin{figure}[pos=t]
\centering

\begin{minipage}[t]{0.485\textwidth}
\centering
\includegraphics[width=\linewidth]{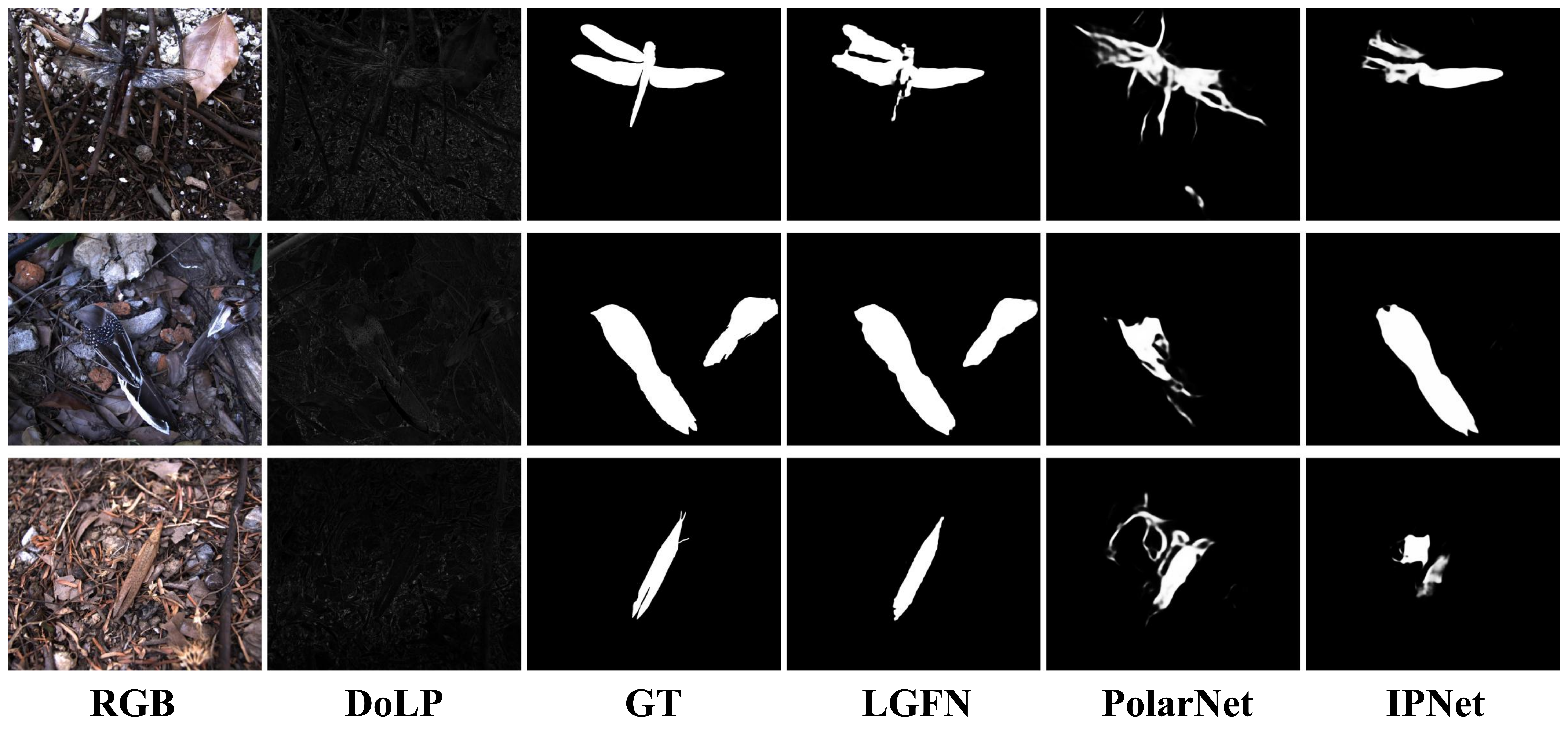}
\end{minipage}
\hfill
\begin{minipage}[t]{0.485\textwidth}
\centering
\includegraphics[width=\linewidth]{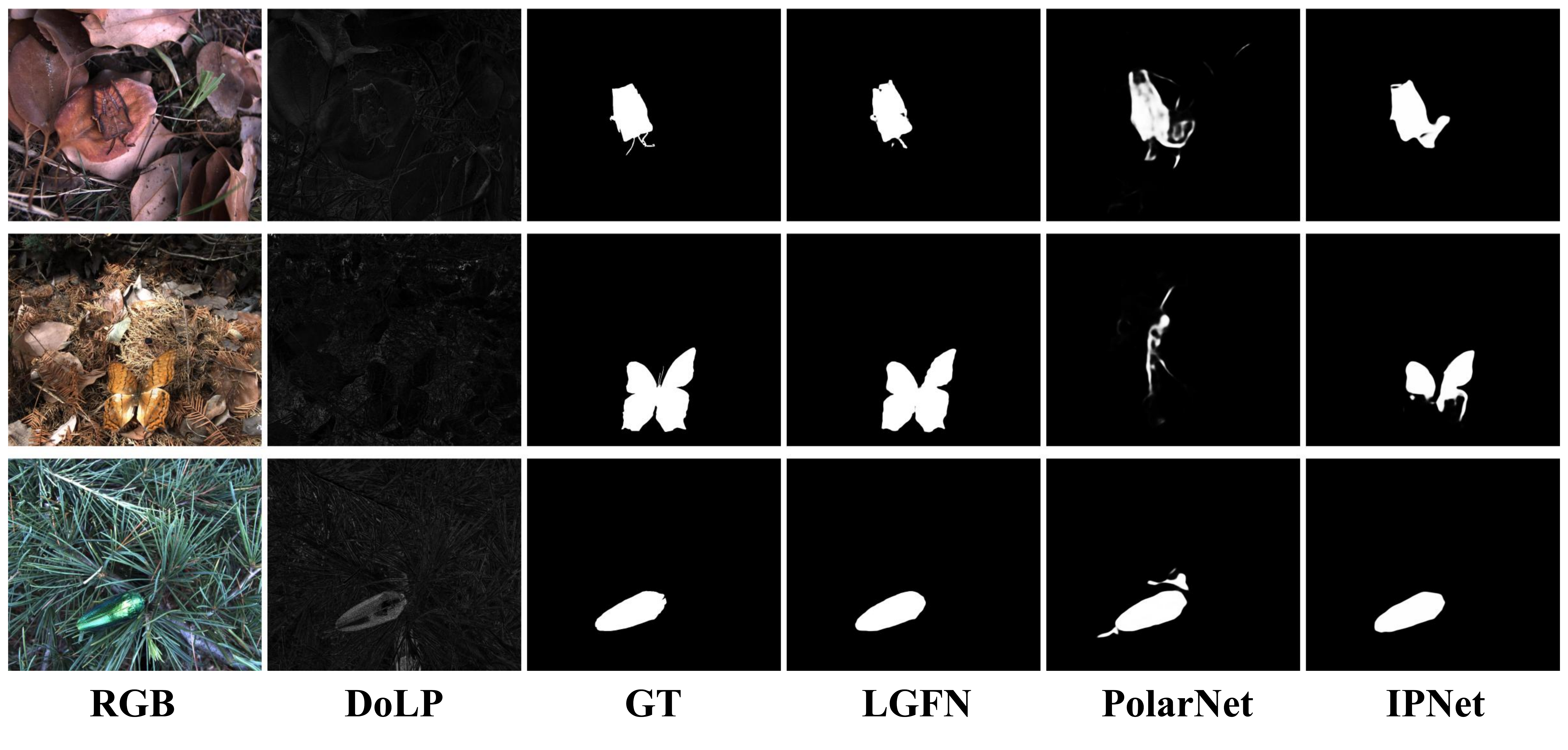}
\end{minipage}

\caption{Qualitative comparison with polarization-assisted camouflaged object detection methods on representative samples from the PCOD\_1200 test set.}
\label{fig:pcod_polar_visual}
\vspace{-0.5em}
\end{figure}
\subsection{Loss Functions}
\label{sec:loss}

With the multimodal representations established by GPH and RPCF, LGFN is optimized using task-level segmentation supervision together with two auxiliary multimodal objectives. The final prediction provides the primary camouflage segmentation supervision, while the RGB and edge branches serve as training-only auxiliary regularizers. In addition, the Fusion Decoder and Modality Gate are constrained by fusion-consistency and modality-allocation objectives, respectively. These auxiliary components are used only during training and introduce no additional inference-time computation.

Following common segmentation practice~\cite{wei2020f3net,milletari2016vnet}, we adopt a structure-aware segmentation objective combining weighted binary cross-entropy, weighted intersection-over-union, and Dice losses:
\begin{equation}
\begin{gathered}
\mathbf{W}
=
1+5\left|
\mathcal{A}_{31}(\mathbf{G})-\mathbf{G}
\right|,
\\[3pt]
\mathcal{L}_{\mathrm{str}}(\mathbf{P},\mathbf{G})
=
\mathcal{L}_{\mathrm{wBCE}}
+
\mathcal{L}_{\mathrm{wIoU}}
+
\mathcal{L}_{\mathrm{Dice}}.
\end{gathered}
\label{eq:structure_loss}
\end{equation}

Here, $\mathbf{P}$ and $\mathbf{G}$ denote the predicted probability map and binary ground truth, respectively, and $\mathcal{A}_{31}(\cdot)$ denotes $31\times31$ average pooling. The weight map $\mathbf{W}$ emphasizes structurally ambiguous boundary regions. Edge supervision $\mathbf{G}_{\mathrm{edge}}$ is generated from the morphological difference between the dilated and eroded ground-truth masks.

The training-only Fusion Decoder regularizes the multimodal representation by preserving RGB-anchored appearance together with complementary polarization structures. Let $\mathbf{P}_{\mathrm{fus}}$ denote its reconstructed fusion map and $\mathbf{I}_{\mathrm{gray}}$ the grayscale RGB image. A gradient target is constructed from the strongest available structural response:
\begin{equation}
\mathbf{T}_{\mathrm{grad}}
=
\max
\left(
|\nabla\mathbf{I}_{\mathrm{gray}}|,
|\nabla\overline{\mathbf{D}}|,
|\nabla\overline{\mathbf{A}}|
\right),
\label{eq:gradient_target}
\end{equation}
where $\overline{\mathbf{D}}$ and $\overline{\mathbf{A}}$ denote the availability-masked DoLP and AoP maps. The fusion-consistency objective is defined as
\begin{equation}
\begin{gathered}
\mathcal{L}_{\mathrm{fusion}}
=
\left\|
\mathbf{P}_{\mathrm{fus}}
-
\mathbf{I}_{\mathrm{gray}}
\right\|_{1}
+
\left\|
\nabla\mathbf{P}_{\mathrm{fus}}
-
\mathbf{T}_{\mathrm{grad}}
\right\|_{1}
\\[3pt]
+
1-
\operatorname{SSIM}
\left(
\mathbf{P}_{\mathrm{fus}},
\mathbf{I}_{\mathrm{gray}}
\right).
\end{gathered}
\label{eq:fusion_loss}
\end{equation}

The intensity and SSIM terms~\cite{wang2004ssim} preserve RGB-anchored appearance, while the gradient term retains complementary structural evidence from the available modalities.

The Modality Gate is regularized using fixed offline targets for each modality-availability pattern. Let $(w_D^{*},w_A^{*},\gamma^{*})$ denote the target polarization allocation and injection strength. The Gate objective is defined as
\begin{equation}
\mathcal{L}_{\mathrm{gate}}
=
\left\|
[w_D,w_A]
-
[w_D^{*},w_A^{*}]
\right\|_{2}^{2}
+
(\gamma-\gamma^{*})^{2}.
\label{eq:gate_loss}
\end{equation}

The offline targets are initialized before training using a 26-D descriptor extracted from the available training modalities and remain fixed throughout optimization. The descriptor is used only for target construction and is never provided to the Modality Gate or any inference pathway.

Let $\mathbf{P}_{\mathrm{final}}$, $\mathbf{P}_{\mathrm{rgb}}$, and $\mathbf{P}_{\mathrm{edge}}$ denote the final prediction, RGB auxiliary prediction, and edge prediction, respectively. The overall training objective is defined as
\begin{equation}
\begin{gathered}
\mathcal{L}_{\mathrm{task}}
=
\mathcal{L}_{\mathrm{str}}
(\mathbf{P}_{\mathrm{final}},\mathbf{G})
+
0.2\mathcal{L}_{\mathrm{str}}
(\mathbf{P}_{\mathrm{rgb}},\mathbf{G})
+
0.5\mathcal{L}_{\mathrm{str}}
(\mathbf{P}_{\mathrm{edge}},\mathbf{G}_{\mathrm{edge}}),
\\[4pt]
\mathcal{L}_{\mathrm{total}}
=
\mathcal{L}_{\mathrm{task}}
+
0.3\mathcal{L}_{\mathrm{fusion}}
+
0.3\mathcal{L}_{\mathrm{gate}}.
\end{gathered}
\label{eq:total_loss}
\end{equation}

The final prediction provides the primary segmentation supervision, while the RGB and edge predictions act as auxiliary training constraints. The fusion-consistency and Gate objectives further regularize multimodal representation learning and availability-conditioned polarization allocation without introducing additional inference-time computation.
\section{Experiments}
\label{sec:experiments}
\begin{table*}[!t]
\centering
\footnotesize
\renewcommand{\arraystretch}{1.0}
{\rmfamily
\caption{Quantitative comparison with polarization-assisted camouflaged object detection methods on the complete 230-image PCOD\_1200 test set under a  adopted local reevaluation protocol. All prediction maps are reevaluated using the same metric implementation. The best and second-best results are highlighted in red and blue, respectively.}
\label{tab:polarization_comparison}

\begin{tabular*}{\textwidth}{
@{\extracolsep{\fill}}
ccccccccc
@{}
}
\toprule
\multirow{2}{*}{Methods}
& \multirow{2}{*}{Year}
& \multirow{2}{*}{Venue}
& \multicolumn{6}{c}{Evaluation metrics} \\
\cmidrule(lr){4-9}
& & &
MAE$\downarrow$
& $S_{\alpha}\uparrow$
& $E_{\phi}\uparrow$
& $F_{\beta}^{w}\uparrow$
& Dice$\uparrow$
& IoU$\uparrow$ \\
\midrule

PolarNet~\cite{wang2023polarnet}
& 2023
& PRL
& 0.0473
& 0.6610
& 0.7718
& 0.4537
& 0.4392
& 0.3403 \\

IPNet~\cite{wang2024ipnet}
& 2024
& EAAI
& \textbf{\color[HTML]{34CDF9}0.0147}
& \textbf{\color[HTML]{34CDF9}0.8696}
& \textbf{\color[HTML]{34CDF9}0.9342}
& \textbf{\color[HTML]{34CDF9}0.8184}
& \textbf{\color[HTML]{34CDF9}0.8063}
& \textbf{\color[HTML]{34CDF9}0.7162} \\

\midrule

\textbf{LGFN }
& 2026
& --
& \textbf{\color[HTML]{FE0000}0.0097}
& \textbf{\color[HTML]{FE0000}0.9045}
& \textbf{\color[HTML]{FE0000}0.9651}
& \textbf{\color[HTML]{FE0000}0.8693}
& \textbf{\color[HTML]{FE0000}0.8678}
& \textbf{\color[HTML]{FE0000}0.7944} \\

\bottomrule
\end{tabular*}
}
\end{table*}

This section evaluates LGFN through comparisons with RGB-based and polarization-assisted COD methods, component and objective ablations, qualitative analysis, and computational-efficiency assessment.

\subsection{Experimental Settings}
\label{sec:experimental_settings}
We conduct experiments on PCOD\_1200~\cite{wang2024ipnet}, COD10K~\cite{fan2020camouflaged}, and NC4K~\cite{lv2021simultaneously}. The complete 230-image test partition of PCOD\_1200 is used for the RGB-based and polarization-assisted comparisons, ablation studies, and qualitative analysis. For additional evaluation on larger RGB benchmarks, we report results on fixed subsets of 405 and 200 images from COD10K and NC4K, respectively. Within each evaluation set, the same image list and metric implementation are used for all compared prediction maps. These evaluations provide complementary evidence of the performance of LGFN across different camouflage benchmarks under a consistent local evaluation setting.

All input images are resized to $352\times352$, and PVT-v2-B2 is adopted as the RGB encoder. The RGB-only and multimodal routes of LGFN are independently optimized and stored as separate checkpoints. The RGB-only route is trained on a mixed RGB training set constructed from the PCOD\_1200 training partition together with fixed training subsets of COD10K and NC4K, whereas the multimodal route is trained for 180 epochs using only the multimodal training partition of PCOD\_1200. The COD10K and NC4K images used for RGB training are strictly disjoint from their corresponding evaluation subsets described above. At inference, the multimodal route receives RGB together with the available DoLP and AoP inputs and the modality-availability vector, without requiring sample-dependent statistics or handcrafted quality descriptors. The two routes therefore represent independently optimized deployment configurations within the overall LGFN design.

For RGB-based evaluation, LGFN with the RGB-only route is compared with nine representative COD methods: PFNet~\cite{mei2021pfnet}, C2FNet~\cite{sun2021c2fnet}, ZoomNet~\cite{pang2022zoomnet}, SINet-V2~\cite{fan2022sinetv2}, FSPNet~\cite{huang2023fspnet}, DGNet~\cite{ji2023dgnet}, VSCode-T~\cite{luo2024vscode}, GBNet~\cite{wang2026gbnet}, and DepthSAM~\cite{han2026depthsam}. For polarization-assisted evaluation, LGFN with the multimodal route is compared with PolarNet~\cite{wang2023polarnet} and IPNet~\cite{wang2024ipnet}. 

We adopt six evaluation metrics: mean absolute error (MAE), structure measure $S_{\alpha}$~\cite{fan2017structure}, enhanced-alignment measure $E_{\phi}$~\cite{fan2018enhanced}, weighted F-measure $F_{\beta}^{w}$~\cite{margolin2014evaluate}, Dice coefficient~\cite{milletari2016vnet}, and intersection over union (IoU). Lower MAE indicates better performance, whereas higher values are preferred for the remaining metrics. For the RGB-based quantitative comparisons, the best, second-best, and third-best results are highlighted in red, blue, and green, respectively. For the polarization-assisted comparison, the best and second-best results are highlighted in red and blue, respectively.

\subsection{RGB-Based Comparison}
\label{sec:rgb_comparison_results}

Tables~\ref{tab:rgb_comparison}, \ref{tab:nc4k_comparison}, and \ref{tab:cod10k_comparison} report the quantitative comparisons with nine RGB-based COD methods, while Figs.~\ref{fig:pcod_rgb_visual}, \ref{fig:nc4k_rgb_visual}, and \ref{fig:cod10k_rgb_visual} present representative qualitative results. On the complete 230-image PCOD\_1200 test set, LGFN with the RGB-only route achieves the best performance across all six metrics. Relative to the best competing values, it reduces MAE from 0.0140 to 0.0090 and improves Dice and IoU from 0.8249 and 0.7714 to 0.8806 and 0.8144, respectively. The concurrent improvements in $S_{\alpha}$, $E_{\phi}$, and $F_{\beta}^{w}$ further indicate stronger structural alignment and foreground recovery. As shown in Fig.~\ref{fig:pcod_rgb_visual}, LGFN produces more complete target masks and suppresses distracting background responses in scenes characterized by weak appearance contrast, fragmented structures, and irregular boundaries.

On the fixed 405-image COD10K evaluation subset, LGFN ranks third in MAE, $S_{\alpha}$, $E_{\phi}$, Dice, and IoU, and fourth in $F_{\beta}^{w}$. Although GBNet and DepthSAM achieve stronger overall results within this evaluation set, LGFN surpasses the remaining compared methods in five of the six metrics. Compared with VSCode-T, it reduces MAE from 0.0308 to 0.0282 and improves $S_{\alpha}$, $E_{\phi}$, Dice, and IoU to 0.8540, 0.9179, 0.7966, and 0.7068, respectively, while yielding a slightly lower $F_{\beta}^{w}$. The qualitative examples in Fig.~\ref{fig:cod10k_rgb_visual} show that LGFN generally preserves the principal target structures; however, fine boundaries and locally ambiguous regions remain challenging.

On the fixed 200-image NC4K evaluation subset, LGFN achieves the best $E_{\phi}$ and Dice scores, ranks second in MAE, $S_{\alpha}$, and IoU, and ranks third in $F_{\beta}^{w}$. Its $E_{\phi}$ of 0.9874 and Dice score of 0.9688 exceed the best competing values, while its MAE, $S_{\alpha}$, and IoU are second only to those of DepthSAM within this evaluation set. This pattern of results suggests a favorable balance among pixel-level accuracy, structural consistency, and region overlap. As illustrated in Fig.~\ref{fig:nc4k_rgb_visual}, the predicted masks are spatially coherent and closely follow the target extent. Together, the COD10K and NC4K subset results provide supplementary evidence that the RGB-only configuration remains competitive under the adopted local evaluation setting.

\begin{table}[!t]
\centering
\footnotesize
\setlength{\tabcolsep}{5.5pt}
\renewcommand{\arraystretch}{1.05}
{\rmfamily
\caption{Ablation study of the core components on the 230-image PCOD\_1200 test set. Each variant is independently trained under the same experimental protocol. The best and second-best results are highlighted in red and blue, respectively.}
\label{tab:module_ablation}

\begin{tabular}{lcccccc}
\toprule
\multirow{2}{*}{Variants}
& \multicolumn{6}{c}{Evaluation metrics} \\
\cmidrule(lr){2-7}
& MAE$\downarrow$
& $S_{\alpha}\uparrow$
& $E_{\phi}\uparrow$
& $F_{\beta}^{w}\uparrow$
& Dice$\uparrow$
& IoU$\uparrow$ \\
\midrule

\textbf{LGFN}
& \textbf{\color[HTML]{FE0000}0.0097}
& \textbf{\color[HTML]{FE0000}0.9045}
& \textbf{\color[HTML]{FE0000}0.9651}
& \textbf{\color[HTML]{FE0000}0.8693}
& \textbf{\color[HTML]{FE0000}0.8678}
& \textbf{\color[HTML]{FE0000}0.7944} \\

\midrule

Baseline (w/o all modules)
& 0.0112
& 0.8975
& 0.9602
& 0.8579
& 0.8559
& 0.7805 \\

w/o RPCF
& \textbf{\color[HTML]{34CDF9}0.0098}
& 0.9009
& \textbf{\color[HTML]{34CDF9}0.9622}
& 0.8589
& 0.8575
& 0.7857 \\

w/o GPH
& 0.0111
& 0.8987
& 0.9573
& 0.8557
& 0.8585
& 0.7811 \\

w/o Gate
& 0.0101
& \textbf{\color[HTML]{34CDF9}0.9016}
& 0.9621
& \textbf{\color[HTML]{34CDF9}0.8656}
& \textbf{\color[HTML]{34CDF9}0.8626}
& \textbf{\color[HTML]{34CDF9}0.7891} \\

w/o Explicit Cues
& 0.0107
& 0.9003
& 0.9581
& 0.8585
& 0.8580
& 0.7837 \\

\bottomrule
\end{tabular}
}
\end{table}
\begin{table}[!t]
\centering
\footnotesize
\setlength{\tabcolsep}{5.5pt}
\renewcommand{\arraystretch}{1.05}
{\rmfamily
\caption{Ablation study of the core training objectives on the 230-image PCOD\_1200 test set. Each variant is independently trained under the same experimental protocol. The best and second-best results are highlighted in red and blue, respectively.}
\label{tab:loss_ablation}

\begin{tabular}{lcccccc}
\toprule
\multirow{2}{*}{Variants}
& \multicolumn{6}{c}{Evaluation metrics} \\
\cmidrule(lr){2-7}
& MAE$\downarrow$
& $S_{\alpha}\uparrow$
& $E_{\phi}\uparrow$
& $F_{\beta}^{w}\uparrow$
& Dice$\uparrow$
& IoU$\uparrow$ \\
\midrule

\textbf{LGFN}
& \textbf{\color[HTML]{FE0000}0.0097}
& \textbf{\color[HTML]{FE0000}0.9045}
& \textbf{\color[HTML]{FE0000}0.9651}
& \textbf{\color[HTML]{FE0000}0.8693}
& \textbf{\color[HTML]{FE0000}0.8678}
& \textbf{\color[HTML]{FE0000}0.7944} \\

\midrule

w/o Gate Loss
& 0.0105
& \textbf{\color[HTML]{34CDF9}0.8993}
& \textbf{\color[HTML]{34CDF9}0.9583}
& \textbf{\color[HTML]{34CDF9}0.8586}
& \textbf{\color[HTML]{34CDF9}0.8561}
& \textbf{\color[HTML]{34CDF9}0.7849} \\

w/o Fusion Loss
& \textbf{\color[HTML]{34CDF9}0.0104}
& 0.8946
& 0.9543
& 0.8520
& 0.8455
& 0.7742 \\

\bottomrule
\end{tabular}
}
\end{table}

\subsection{Polarization-Assisted Comparison}
\label{sec:polarization_comparison_results}

We next compare LGFN with the polarization-assisted PolarNet and IPNet on the complete 230-image PCOD\_1200 test set. In this evaluation, LGFN operates through its multimodal route. As reported in Table~\ref{tab:polarization_comparison}, LGFN achieves the best performance across all six metrics under the adopted local reevaluation protocol. Compared with IPNet, it reduces MAE from 0.0147 to 0.0097 and improves $S_{\alpha}$, $E_{\phi}$, and $F_{\beta}^{w}$ by 0.0349, 0.0309, and 0.0509, respectively. The gains are particularly pronounced in region overlap, with Dice increasing from 0.8063 to 0.8678 and IoU from 0.7162 to 0.7944. These results indicate improved foreground localization, structural consistency, and mask completeness.

The qualitative comparison in Fig.~\ref{fig:pcod_polar_visual} is consistent with the quantitative results. PolarNet frequently produces weak or fragmented foreground responses, whereas IPNet generally recovers the principal target regions but may retain internal omissions and background interference. In contrast, LGFN produces more complete object structures and cleaner boundaries, particularly when foreground and background exhibit highly similar appearance characteristics. This behavior reflects the proposed information flow: GPH first coordinates heterogeneous DoLP, AoP, and explicit polarization evidence, after which RPCF introduces the resulting representation into the RGB hierarchy through calibrated residual interaction.

Together, the quantitative and qualitative results support the effectiveness of decoupling polarization-domain coordination from RGB--polarization interaction under the adopted local reevaluation protocol.

\subsection{Ablation Studies}
\label{sec:ablation_studies}

We conduct component and objective ablations on the complete 230-image PCOD\_1200 test set. Each variant is independently trained for 180 epochs under the same protocol, with only the examined component or objective removed.

As shown in Table~\ref{tab:module_ablation}, removing all proposed components degrades all six metrics, with MAE increasing from 0.0097 to 0.0112 and IoU decreasing from 0.7944 to 0.7805. Among the individual variants, removing GPH leads to the most pronounced overall degradation, highlighting the importance of intra-polarization coordination. Removing RPCF or the explicit cues also consistently reduces performance, demonstrating the complementary value of controlled cross-modal interaction and source-level structural information. The Modality Gate yields a smaller but consistent gain, supporting its role in availability-conditioned polarization calibration.

Table~\ref{tab:loss_ablation} shows that both auxiliary multimodal objectives contribute to the final performance. Removing the Fusion loss reduces $F_{\beta}^{w}$, Dice, and IoU by 0.0173, 0.0223, and 0.0202, respectively, while removing the Gate loss decreases Dice and IoU to 0.8561 and 0.7849. These results indicate that fusion-consistency supervision and availability-conditioned polarization allocation provide complementary regularization during multimodal training without introducing additional inference-time computation.

\subsection{Computational Efficiency}
\label{sec:efficiency_analysis}

We evaluate computational efficiency on an NVIDIA A800-SXM4-80GB GPU using an input resolution of $352\times352$ and a batch size of one. After 30 warm-up iterations, latency is measured over 200 synchronized forward passes, and the procedure is repeated five times to obtain the reported average. The measured latency includes network forward propagation only, while floating-point operations (FLOPs) are estimated using THOP. Since theoretical operation counts depend on operator implementation and profiling support, they are reported together with wall-clock latency rather than treated as a direct measure of practical speed. All complexity results are reported for the checkpoint executed in the corresponding configuration. The reported latency excludes checkpoint loading, preprocessing, and post-processing.

As reported in Table~\ref{tab:rgb_efficiency}, LGFN with the RGB-only route has the lowest measured inference latency among the evaluated models, requiring 7.40~ms per image, while requiring 25.1004~M parameters and 10.3591~G FLOPs, ranking second in both model size and computational cost. Although DGNet has fewer parameters and lower reported FLOPs, its measured latency is higher, illustrating that theoretical complexity does not necessarily translate directly into runtime performance. Together with its strong segmentation performance, these results indicate a favorable balance among accuracy, model compactness, and practical inference speed.
\begin{figure}[pos=t]
\centering

\begin{minipage}[t]{0.485\textwidth}
\centering
\includegraphics[width=\linewidth]{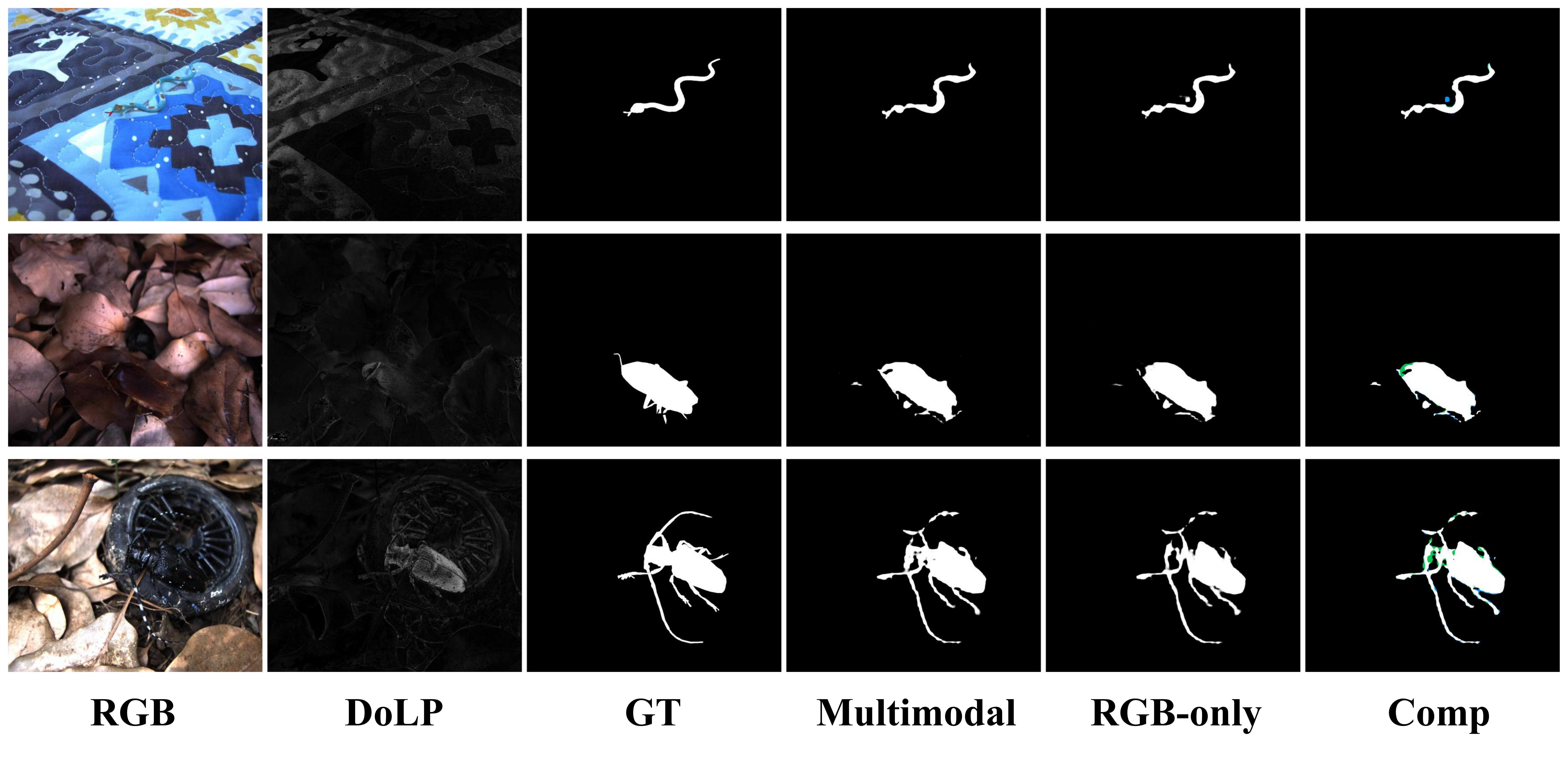}
\end{minipage}
\hfill
\begin{minipage}[t]{0.485\textwidth}
\centering
\includegraphics[width=\linewidth]{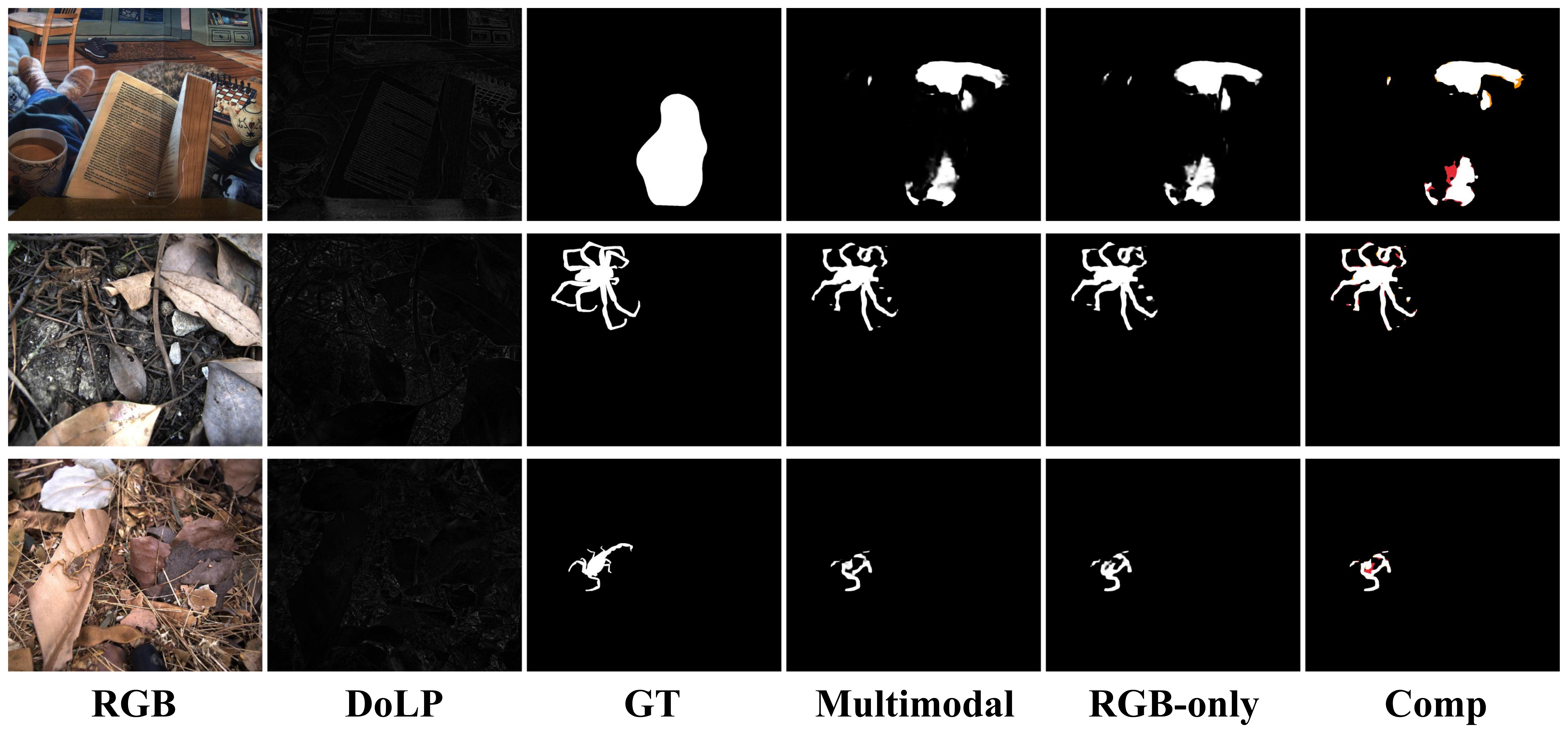}
\end{minipage}

\caption{Qualitative analysis of polarization assistance on the PCOD\_1200 test set. Left: successful cases; right: failure cases. Green and cyan denote recovered targets and removed false positives, whereas red and orange denote additional false negatives and false positives, respectively.}
\label{fig:polarization_case_analysis}
\end{figure}
\begin{table}[!t]
\centering
\footnotesize
\setlength{\tabcolsep}{6.0pt}
\renewcommand{\arraystretch}{1.05}
{\rmfamily
\caption{Computational efficiency comparison with RGB-based camouflaged object detection methods. The best and second-best results are highlighted in red and blue, respectively.}
\label{tab:rgb_efficiency}

\begin{tabular}{lccccc}
\toprule
\multirow{2}{*}{Methods}
& \multirow{2}{*}{Year}
& \multirow{2}{*}{Venue}
& \multicolumn{3}{c}{Computational efficiency} \\
\cmidrule(lr){4-6}
& & &
Params (M)$\downarrow$
& FLOPs (G)$\downarrow$
& Time (ms)$\downarrow$ \\
\midrule

PFNet~\cite{mei2021pfnet}
& 2021
& CVPR
& 46.4978
& 19.0444
& \textbf{\color[HTML]{34CDF9}8.7004} \\

C2FNet~\cite{sun2021c2fnet}
& 2021
& IJCAI
& 28.4112
& 13.1565
& 10.7227 \\

ZoomNet~\cite{pang2022zoomnet}
& 2022
& CVPR
& 32.3815
& 86.2405
& 16.7449 \\

SINet-V2~\cite{fan2022sinetv2}
& 2022
& TPAMI
& 26.9756
& 12.3135
& 9.3346 \\

FSPNet~\cite{huang2023fspnet}
& 2023
& CVPR
& 274.1693
& 238.0916
& 23.7252 \\

DGNet~\cite{ji2023dgnet}
& 2023
& MIR
& \textbf{\color[HTML]{FE0000}19.2235}
& \textbf{\color[HTML]{FE0000}2.8658}
& 11.9622 \\

VSCode-T~\cite{luo2024vscode}
& 2024
& CVPR
& 54.1137
& 61.3283
& 30.9950 \\

GBNet~\cite{wang2026gbnet}
& 2026
& TIP
& 77.1296
& 35.5363
& 24.2087 \\

DepthSAM~\cite{han2026depthsam}
& 2026
& CVPR
& 362.8366
& 379.7616
& 125.2875 \\

\midrule

\textbf{LGFN}
& 2026
& --
& \textbf{\color[HTML]{34CDF9}25.1004}
& \textbf{\color[HTML]{34CDF9}10.3591}
& \textbf{\color[HTML]{FE0000}7.4031} \\

\bottomrule
\end{tabular}
}
\end{table}

Table~\ref{tab:multimodal_efficiency} reports the corresponding comparison among polarization-assisted models. With the multimodal route, LGFN requires 59.2043~M parameters, 24.2923~G FLOPs, and 18.4563~ms per image. Compared with IPNet, it reduces the parameter count, FLOPs, and latency by 53.1\%, 73.6\%, and 63.0\%, respectively, while performing better on all six segmentation metrics. Although PolarNet is lighter and faster, LGFN achieves substantially stronger segmentation performance while maintaining moderate computational cost.

\begin{table}[!t]
\centering
\footnotesize
\setlength{\tabcolsep}{7.0pt}
\renewcommand{\arraystretch}{1.05}
{\rmfamily
\caption{Computational efficiency comparison with polarization-assisted camouflaged object detection methods. The best and second-best results are highlighted in red and blue, respectively.}
\label{tab:multimodal_efficiency}

\begin{tabular}{lccccc}
\toprule
\multirow{2}{*}{Methods}
& \multirow{2}{*}{Year}
& \multirow{2}{*}{Venue}
& \multicolumn{3}{c}{Computational efficiency} \\
\cmidrule(lr){4-6}
& & &
Params (M)$\downarrow$
& FLOPs (G)$\downarrow$
& Time (ms)$\downarrow$ \\
\midrule

PolarNet~\cite{wang2023polarnet}
& 2023
& PRL
& \textbf{\color[HTML]{FE0000}27.6478}
& \textbf{\color[HTML]{FE0000}7.9311}
& \textbf{\color[HTML]{FE0000}9.7608} \\

IPNet~\cite{wang2024ipnet}
& 2024
& EAAI
& 126.1518
& 92.1585
& 49.8998 \\

\midrule

\textbf{LGFN}
& 2026
& --
& \textbf{\color[HTML]{34CDF9}59.2043}
& \textbf{\color[HTML]{34CDF9}24.2923}
& \textbf{\color[HTML]{34CDF9}18.4563} \\

\bottomrule
\end{tabular}
}
\end{table}

\subsection{Qualitative Analysis}
\label{sec:qualitative_analysis}

Figure~\ref{fig:polarization_case_analysis} qualitatively compares the independently optimized RGB-only and multimodal routes of LGFN on the PCOD\_1200 test set. In the successful cases, the multimodal route recovers weak or thin target regions missed by the RGB-only route and suppresses distracting background responses. These observations indicate that polarization-assisted representation can provide complementary physical evidence when RGB appearance alone is ambiguous.

In more challenging cases, strong background polarization responses, spatially inconsistent target cues, or conflicts between RGB and polarization evidence may introduce additional false negatives or false positives, as indicated by the red and orange regions in Fig.~\ref{fig:polarization_case_analysis}. Since the two routes are independently optimized for different input configurations, their prediction differences reflect both polarization-assisted representation and route-specific optimization. These cases motivate finer sample- and region-aware polarization reliability modeling in future work.

\section{Conclusion}
\label{sec:conclusion}

This paper presented LGFN, a lightweight gated RGB--polarization fusion framework for camouflaged object detection under different sensor-input configurations. A deterministic Modality Router selects a separately optimized RGB-only or multimodal configuration according to polarization availability. In the multimodal configuration, the Modality Gate calibrates the available polarization branches, GPH coordinates learned DoLP and AoP representations with explicit structural cues, and RPCF introduces the coordinated polarization representation into the RGB hierarchy through controlled residual interaction. This design separates intra-polarization coordination from RGB--polarization interaction while retaining RGB as the principal representation pathway.

On the complete PCOD\_1200 test set, the RGB-only configuration achieves the best results on all six metrics among the evaluated RGB-based methods and remains competitive on the fixed COD10K and NC4K evaluation subsets. The multimodal configuration outperforms PolarNet and IPNet under the adopted local reevaluation protocol. Component and objective ablations further support the contributions of GPH, RPCF, the Modality Gate, and the auxiliary training objectives. The current study is limited to a single polarization-based benchmark and uses separately optimized checkpoints for the two deployment configurations. Future work will investigate shared parameterization and finer sample- and region-aware polarization reliability modeling under more diverse sensing conditions.

\section{Acknowledgements}
This research was supported by the Basic and Applied Basic Research of Guangdong Province (No. 2023A1515140077), the Natural Science Foundation of Guangdong Province (No. 2024A1515011880), the Research Fund of Guangdong-HongKong-Macao Joint Laboratory for Intelligent Micro-Nano Optoelectronic Technology (No. 2020B1212030010).

\printcredits
\bibliographystyle{cas-model2-names}

\bibliography{references}


\end{document}